\documentclass{article}

\usepackage{PRIMEarxiv}

\usepackage[utf8]{inputenc} 
\usepackage[T1]{fontenc}    
\usepackage{hyperref}       
\usepackage{url}            
\usepackage{booktabs}       
\usepackage{amsfonts}       
\usepackage{nicefrac}       
\usepackage{microtype}      
\usepackage{lipsum}
\usepackage{fancyhdr}       
\usepackage{graphicx}       
\graphicspath{{media/}}     
\usepackage{amsmath}

\title{Double-stage Feature-level Clustering based Mixture of Experts Framework
}

\author{
  Bakary Badjie, José Cecílio and António Casimiro \\
  LASIGE, Departamento de Informática \\
  Faculdade de Ciências da Universidade de Lisboa \\
  Lisboa\\
  \texttt{\{bbadjie, jmcecilio, casim\}@ciencias.ulisboa.pt}
}

\begin{document}
\maketitle

\begin{abstract}

The Mixture-of-Experts (MoE) model has succeeded in deep learning (DL). However, its complex architecture and advantages over dense models in image classification remain unclear. In previous studies, MoE performance has often been affected by noise and outliers in the input space. Some approaches incorporate input clustering for training MoE models, but most clustering algorithms lack access to labeled data, limiting their effectiveness.
This paper introduces the Double-stage Feature-level Clustering and Pseudo-labeling-based Mixture of Experts (DFCP-MoE) framework, which consists of input feature extraction, feature-level clustering, and a computationally efficient pseudo-labeling strategy. This approach reduces the impact of noise and outliers while leveraging a small subset of labeled data to label a large portion of unlabeled inputs. We propose a conditional end-to-end joint training method that improves expert specialization by training the MoE model on well-labeled, clustered inputs.
Unlike traditional MoE and dense models, the DFCP-MoE framework effectively captures input space diversity, leading to competitive inference results. We validate our approach on three benchmark datasets for multi-class classification tasks.

\end{abstract}


\sloppy

\section{Introduction}

Image classification is essential in many real-world applications, such as autonomous driving systems (ADS) and medical imaging (e.g., radiography). However, the vast amount of data in the physical world presents a challenge for conventional deep learning (DL) methods like deep neural networks (DNNs), which typically train on entire datasets. DNNs struggle to capture the full complexity of large-scale data without significantly increasing computational demands. Furthermore, scaling dense models to accommodate massive datasets during training is impractical, as it can lead to increased latency during inference, negatively impacting performance \cite{wang2024toward}.

Ensemble learning has been introduced to address these scalability challenges by combining multiple models to improve predictive accuracy while reducing inference latency. Traditional ensemble methods train various models on the same dataset and aggregate their predictions using techniques like bagging (majority voting), boosting (error correction), or stacking (meta-learning) ~\cite{antonucci2013ensemble, cheng2010bayes}. While these approaches enhance predictive precision compared to single-model training, they have limitations. Each model contributes to every inference, which reduces flexibility in selecting the most suitable models, increases computational overhead, and limits scalability due to high memory and processing requirements. Additionally, ensemble methods struggle with multi-modal data (e.g., integrating text, images, and speech) because models are often optimized for a single feature space \cite{odegua2019empirical, kim2022learning}.

To overcome these challenges, we propose the Double-stage Feature-level Clustering and Pseudo-labeling-based Mixture of Experts (DFCP-MoE) framework, an improved Mixture-of-Experts (MoE)~\cite{jordan1994hierarchical} approach for image classification. Traditionally, MoE utilizes multiple small-scale neural networks (experts) and a managing network (gating network) to handle large and diverse datasets. The managing network determines which experts process each input, optimizing both training and inference efficiency~\cite{vats2024evolution}.

MoE models are typically trained using one of two methodologies: (1) gradient descent with backpropagation~\cite{jacobs1991task}, where the managing network assigns weights to experts based on training data, and (2) Expectation-Maximization (EM)~\cite{jordan1994hierarchical}, which iteratively refines expert specialization.
In this work, we employ gradient descent, where each expert transforms an input into an output, and the overall MoE output is a weighted sum of expert outputs. The managing network assigns these weights using a softmax function, ensuring that experts specialize in different areas of the input space. Unlike traditional ensemble methods, MoE allows selective expert activation during inference, reducing computational costs while maintaining performance~\cite{haykin1998neural}.

Most existing MoE architectures rely on random data assignment to experts, entirely depending on the gating mechanism to determine expert suitability, which leads to several issues:
\begin{itemize}
    \item Shrinking batch problem: If the managing network selects a subset of experts per input, each expert receives significantly fewer training samples, limiting learning effectiveness.

    \item Bias in expert selection: Early-selected experts tend to dominate training, leading to imbalanced learning and poor generalization.
    
\end{itemize}

To address these limitations, the DFCP-MoE framework introduces:
\begin{itemize}
    \item Feature-level clustering: The input space is divided into clusters based on feature similarity.
    \item Pseudo-labeling: A small subset of labeled data is used to label a larger proportion of unlabeled inputs.
    \item Conditional expert training: Instead of random assignment, the managing network directs specific clusters to specific experts, improving specialization. Experts train on their assigned clusters while also receiving a small portion of samples from other clusters to prevent overfitting.
\end{itemize}

This structured input assignment allows experts to capture the statistical patterns of their designated class distributions while maintaining adaptability. Additionally, the managing network remains trained on data distribution, ensuring robust decision-making during inference.
Unlike traditional ensemble and dense models, the DFCP-MoE approach optimally assigns computational resources, reduces inference latency, and improves expert specialization. We validate the effectiveness of DFCP-MoE on benchmark datasets, demonstrating its superiority in multi-class classification. The DFCP-MoE approach effectively enhances the MoE model’s ability to handle large-scale image classification tasks.

The rest of this paper is organized as follows. Section~\ref{sec:litera} reviews the contribution of previous works related to MoE and discusses how these contributions are improved through the DFCP-MoE framework that we present in this paper. Section~\ref{sec:bak&contx} presents the proposed DFCP-MoE framework, emphasizing the use of double-stage feature-level clustering and pseudo-labeling approaches and the structure of the MoE architecture. 
Section~\ref{Sec:resuldis} provides experimental results and discusses the observed improvements achieved with the DFCP-MoE framework. Finally, Section~\ref{Sec:Conlfuwork} provides the conclusions and delineates the direction for further research.


 \section{Related Work}
\label{sec:litera}

Several methods have been proposed to develop machine learning (ML) models capable of handling complex real-world scenarios with high predictive accuracy and low latency~\cite{brown2020language, lepikhin2020gshard, raffel2020exploring}. However, optimizing ML models to efficiently adapt and specialize in diverse large-scale datasets without compromising precision and latency remains an open research challenge~\cite{masoudnia2014mixture, das2022comparison}. This challenge introduced the Mixture-of-Experts (MoE) model~\cite{jordan1994hierarchical, jacobs1991adaptive}.
Initially, MoE frameworks employed traditional ML models as expert components, including decision trees~\cite{peralta2014embedded}, support vector machines (SVMs)\cite{tresp2000mixtures}, and hidden Markov models~\cite{jordan1996hidden}. However, these models' representational capacity and conventional learning mechanisms impose significant limitations, restricting their ability to effectively manage complex and diverse data distributions.

Eigen et al.~\cite{Eigen2013LearningFR} introduced a multi-layered MoE model consisting of several layers of experts and routers, improving the MoE's ability to handle intricate datasets. Complementarily, Shazeer et al.~\cite{shazeer2016outrageously} further enhanced the performance of the MoE model by introducing sparsity to the router's output for each training sample. This modification enhances training consistency and decreases computing overhead. 
Furthermore, Hazimeh et al. \cite{hazimeh2021dselect} developed the "DSelect-k" method, a continuously differentiable "sparse gate" for MoE. They trained the gate using a first-order approach, such as stochastic gradient descent, to provide explicit expert selection control.

Chen et al.~\cite{chen2022towards} proposed an MoE model incorporating inherent cluster topologies. While their approach designs the router to capture cluster-centered features, the predefined allocation of clusters to expert models may limit its flexibility in making adaptive assignments. This constraint could hinder the model’s ability to adjust to complex statistical distributions during inference. Additionally, their method does not explicitly address the impact of outliers, which may negatively affect the predictive performance of the MoE model.

The DFCP-MoE framework leverages a double-stage clustering approach to refine cluster assignments, overcoming the limitations of previous methods. By dynamically optimizing cluster allocation during training, we jointly fine-tune hyperparameters and the MoE model, allowing the gating network to assign expert weights based on input feature characteristics. This end-to-end conditional training enhances predictive accuracy while significantly reducing inference latency, achieving state-of-the-art performance in image classification tasks.



 \section{Double-stage Feature-level Clustering and Pseudo-labeling-based Mixture of Experts}
\label{sec:bak&contx}

The Double-stage Feature-level Clustering and Pseudo-labeling-based Mixture of Experts (DFCP-MoE) framework is composed of \textit{N} expert models, represented as \( E = E_1, E_2, E_3, \dots, E_N \), along with a managing network \textit{(G)}, all of which are trained together across \textit{K} clusters. Each cluster captures a specific class distribution and consists of \textit{S} labeled training samples that collectively reflect the original data distribution \textit{D}.  

During joint training, the managing network refines its ability to assess each expert's strengths by learning from their errors. This process enhances its capability to allocate appropriate weights to experts based on the input. At inference time, it dynamically distributes weights among experts, determining their contribution to the final prediction. Each expert produces a binary output, indicating whether the input belongs to its assigned class distribution. The model prediction is obtained by computing the gate-weighted sum of all expert outputs.

To compute the gating coefficients (gating weights), a softmax function is applied to the output of the managing network \textit{G}. These coefficients, determined by the managing network, are denoted as \( \pi_{j} (X_i;\Theta_g) \), where \( X_i \) represents the input point, and \( \Theta_g \) corresponds to the parameters of the managing network. The softmax function ensures that \( \pi_{j} (X_i;\Theta_g) \) are non-negative and sum to 1 across all experts. The mathematical formulation of the softmax function within the MoE model is provided in Equation~\ref{eq-1}.

\begin{equation} \label{eq-1}
 \pi_{j} = softmax(X_i; \Theta_g) = \frac{\exp(h_j (X_i; \Theta_g))}{\sum_{j' =1}^E \exp(h_{j'}(X_i; \Theta_g))}  
\end{equation}

In this equation, \( h_j(X_i; \Theta_g) \) represents the output of the gating function, where \( \Theta_g \) is the unique parameter associated with the \textit{j-th} expert \( E_j \) for a given input \( X_i \). The total number of experts in the MoE model is denoted by \( E \), while \( j \) refers to the index of a specific expert. The term \( \pi_{j} \) represents the softmax probability assigned to the \textit{j-th} expert for the input \( X_i \).

During the training phase of the DFCP-MoE framework, the managing network adaptively computes softmax-driven gating weights based on input characteristics while being guided by cluster-specific training samples. In the inference stage, the managing network utilizes the gating function \( G(X_i) \) to generate a set of weights \( g_j(X_i) \), forming a probability distribution assigned to the relevant experts for a given input. The total sum of these weights across all experts for each input is always equal to \( 1 \), as mathematically defined in Equation~\ref{eq-gateProba}.

\begin{equation} \label{eq-gateProba}
        \sum_{j=1}^E g_j(X_i)  = 1
\end{equation}

Let \( P(Y|X_i, E_j) \) represent the probability distribution of the predicted class \textit{Y} for a given input \( X_i \) and expert \( E_j \). Since the managing network assigns softmax-driven weights \( g_j(X_i) \) to each expert, the contribution of an expert is directly proportional to its assigned weight, mathematically expressed as \( g_j(X_i) = P(E_j|X_i) \). This means that the weight magnitude reflects the managing network’s confidence in that expert for a given input during inference. Consequently, the DFCP-MoE framework's final output for \( X_i \) is determined as follows.

\begin{equation} \label{eq-out}
    \begin{split}
        P(Y|X_i) = \sum_{j=1}^E P(E_j|X_i)P(Y|X_i,E_j) \\ = \sum_{j=1}^E g_j(X_i)P(Y|X_i,E_=j)   
    \end{split}
\end{equation}

\noindent where $P(E_j|X_i)$ is the probability of the \textit{j-th} expert given input $X_i$.

The gating function \( G(X_i) \) dynamically selects among \textit{N} experts based on different clusters, introducing diversity into the DFCP-MoE framework during training. This adaptability enhances the model's overall precision during inference. The general formulation of the DFCP-MoE framework is given as follows.

\begin{equation}\label{eq-3}
    MoE(X_i; \Theta) = \sum_{j = 1}^K G(X_i; \Theta_g)_j \cdot E_j(X_i; \Theta_e)
\end{equation}

\noindent where $G(X_i; \Theta_g)_j$ and $E_j(X_i; \Theta_e)$ are the functions for the managing network and \textit{j-th} expert model with respect to their active parameters $\Theta_g$ and $\Theta_e$, respectively, and a given input $X_i$.

The DFCP-MoE framework employs a deep convolutional neural network (DCNN) as the managing network. It delineates the input-output mapping by correlating the input \( X_i \) to a multidimensional latent space. Equation~\ref{eq-7} specifies this mapping.

\begin{equation}\label{eq-7}
    h(X_i;\Theta_g) = \sum_{p \in [P]} \Theta_g^T X_i^{(P)}
\end{equation}

This mapping consists of parameters $[\Theta_{g_1}, \Theta_{g_2}, \Theta_{g_3},.....\Theta_{g_K}] \in \mathbb{R}^{d \times N}$, which correspond to weights for computing the gating logit for \textit{N} number of experts. $X_i^{(P)}$ denotes the $P^{th}$ division of the input space.

The managing network generates a set of gating logits (weights) equal to the unnormalized scores associated with \( E_j \) for a given input. These logits are then processed through a softmax function, normalizing them into a probability vector distributed across all experts. This is mathematically represented in Equation~\ref{eq-1}.

The DFCP-MoE framework employs two-layer nonhomogeneous DCNNs as expert models, where each expert has a unique number of neurons in its fully connected layers, along with distinct hyperparameters. The specific values of these parameters and hyperparameters for each expert are detailed in Table~\ref{tab:hyper}. The mathematical formulation for the \textit{j-th} expert model \( E_j \) is provided in Equation~\ref{eq-8}.

\begin{equation}\label{eq-8}
    E_j(X, \Theta_e) = \sum_{s \in [F_l]} \sum_{p=1}^P \sigma(\langle \Theta_{s,j}, X^{(p)} \rangle)
\end{equation}

Here $\Theta_{s,j} \in \mathbb{R}^d$ is the weight vector of the \textit{j-th} filter in the \textit{j-th} expert; $F_l$ is the number of filters, \textit{P} is the number of input features, $\Theta_{s,j}$ represent the parameters, $\sigma$ represent an activation function, and $\langle \Theta_{s,j}, X^{(p)} \rangle$ represent the dot product connecting the weights of the \textit{j-th} expert and the \textit{p-th} feature in the input space.

\subsection{Data Distribution and Preparation}
\label{sec:data-dis}

In this experiment, we split the training data into labeled and unlabeled sets. Unlabeled samples are extracted, followed by a feature-level clustering process. The clustered features are reconstructed into images, and pseudo-labeling is applied. Further details are provided in subsequent subsections.


\subsubsection{Feature Extraction}
\label{sec:fea-Ex}

During feature extraction, an EfficientNet-B1 pre-trained CNN model (\url{https://pytorch.org/vision/main/models/generated/torchvision.models.efficientnet_b1.html}) is utilized to extract meaningful features from unlabeled images. Since pre-trained CNN models are extensively trained on large datasets like ImageNet, they effectively capture both low- and high-level feature representations. To focus solely on feature extraction rather than classification, the network is modified by removing the final classification layer, allowing the remaining layers to function as feature extractors. As a result, the model outputs a feature vector instead of class probabilities.

Consider $f_{NN}(\cdot)$ as a pre-trained CNN model, and let $X_i$ represent the image sample of dimension $height (H)$, $width (W)$, and $channel (C)$. After eliminating its last layer, we denote the pre-trained network as $f_{NN}^*(\cdot)$. During each forward pass, $f_{NN}^*(\cdot)$ maps $X_i$ to a high-dimensional feature vector representation. The forward pass is described as follows:

\begin{equation}\label{eq:fow-pass}
    Z_i = f_{NN}^*(X_i)
\end{equation}

\noindent where $Z_i \in \mathbb{R}^d$ is the feature vector representation of dimension $d$.

In the forward pass, the feature extraction required a convolutional operation for the network to generate a feature map or vector. The extraction operation is defined as follows:

\begin{equation}\label{eq:fea-maps}
        Z_i^k (a, q) = \sum_{c=1}^C \sum_{f_a=1}^F  \sum_{f_q=1}^F X_i(a + f_a, q + f_q, c) \cdot 
        L_k(f_a, f_q, c) + b_k
\end{equation}

Here, \( Z_i^k \) represents the output feature map, while \( L_k \) denotes the learned filter. The dimensions of the \textit{K-th} pre-trained filter are given by \( F \) and \( C \), where \( F \) corresponds to the filter size and \( C \) represents the depth (color channels). The term \( (a, q) \) indicates the spatial location within the acquired feature map. The summation over \( c \) iterates through the color channels of \( X_i \), while the summations over \( f_a \) and \( f_q \) traverse the height and width of the \textit{K-th} filter. To maintain the spatial integrity of the extracted features, no pooling operation is applied. This ensures that the extracted features retain essential spatial attributes, leading to higher-quality clustered feature reconstruction when converting them back into images.

\subsubsection{Data Clustering (First Stage Clustering)}
\label{sec:int-Cl}

The traditional K-means clustering algorithm~\cite{lloyd1982least} is employed to group extracted features into \textit{K} initial clusters. It minimizes the objective function \( J = \sum_{i = 1}^K \sum_{Z_i\in C_i} \|Z_i -\mu_i\|^2 \), where \( Z_i \) represents the \textit{i-th} feature map in cluster \( C_i \), and \( \mu_i \) is the centroid (mean feature space) of \( C_i \). This minimization ensures optimal clustering by defining precise, well-represented, and evenly distributed clusters in the input space.

Let the acquired initial clusters be denoted as $C_i={C_{i_1}, C_{i_2}, C_{i_3},...., C_{i_K}}$, where the centroid $\mu_i$ of each initial cluster is computed individually. The centroid of \textit{j-th} initial cluster is computed as $\mu_i = \frac{1}{|C_j|} \sum_{Z\in C_j}Z_i$, where $|C_j|$ and $\sum_{Z\in C_j}Z_i$ represent a set of feature maps and the sum of all the feature maps in the \textit{j-th} initial cluster, respectively.

The \textit{Davies-Bouldin Index (DBI)} score~\cite{singh2020clustering} is calculated to evaluate intra-cluster similarity (within clusters) and inter-cluster similarity (between clusters). This metric ensures that clusters are well-separated in the input space and that data points within each cluster are statistically similar. The DBI is mathematically defined as follows:

\begin{equation} \label{eq:metric}
DBI = \frac{1}{K} \sum_{j=1}^K \underset{u \neq v}{max} (\frac{d_u + d_v}{||\mu_u, \mu_v||})
\end{equation}

In Equation~\ref{eq:metric}, \( K \) denotes the number of clusters, while \( d_u \) and \( d_v \) represent the mean distance between each feature map and the centroids of clusters \( u \) and \( v \), respectively. The term \( \underset{u \neq v}{max} \) indicates the maximum ratio for cluster \( u \) relative to all other clusters, and \( ||\mu_u, \mu_v|| \) represents the distance between the centroids of clusters \( u \) and \( v \). The initial clustering resulted in a higher DBI score (\( > 1 \)), suggesting the clustering process was inaccurate and poorly representative. Ideally, a DBI score of \( 0 \) indicates perfect clustering, scores between \( 0 \) and \( 1 \) are considered acceptable, and scores \( > 1 \) indicate poor clustering. A score closer to \( 0 \) signifies a better clustering solution compared to a score closer to \( 1 \).

In certain scenarios, the complex structure of the data distribution and its high-dimensional features can hinder the effectiveness of a single standard clustering step in producing well-defined clusters~\cite{yu2015incremental}. As a result, a second-stage clustering step is often required to ensure that the final clusters contain only informative features of the inputs. This additional step helps reduce noise and outliers, thereby improving the overall quality and precision of the clustering process.

\subsubsection{Second-stage Clustering}
\label{sec:seco-CL}

A second clustering step is introduced to address the high DBI score issue identified in the first clustering stage. This step refines the initial clusters by focusing on their proximity to neighboring clusters. Using the "K-nearest neighbor" rule, clusters with three or more nearest neighbors (\( K \geq 3 \)) relative to their centroid \( \mu_i \) are selected. Each chosen cluster serves as a reference cluster \( C_{i_f} \), and feature maps from neighboring clusters are selected based on the Euclidean distance metric to form new, refined clusters. A cluster is considered a neighbor in this refinement process if it is sufficiently close to the centroid of the reference cluster (new cluster).

For each feature map in every initial cluster \( C_i \), the Euclidean distance to the centroid of the reference cluster \( \mu_{i_f} \) is calculated as \( d(Z_i, \mu_{i_f}) = \|Z_i - \mu_{i_f}\| \). If \( d(Z_i, \mu_{i_f}) \leq 0.8 \), the feature map \( Z_i \) is assigned to a new cluster \( C_n \). The threshold of \( 0.8 \) is chosen to accurately select the most relevant input features from nearby clusters for forming new clusters. To determine this ideal threshold, an empirical experiment is conducted, testing a range of potential values (from \( 0.1 \) to \( 0.9 \)) and evaluating cluster quality. This threshold ensures that the resulting clusters are well-defined and meaningful.

Unfortunately, many current unsupervised learning methods struggle to align features across domains or generalize to new data points in complex data distributions. This limitation arises because they lack access to labeled data and cannot effectively identify discriminative features, especially in noisy datasets~\cite{bekker2020learning}. To address this, we introduce a pseudo-labeling approach that leverages knowledge from labeled data to annotate data points within a set of newly formed clusters \( C_N = \{C_{n_1}, C_{n_2}, \dots, C_{n_K}\} \).

\subsubsection{Data Labeling (Pseudo-labeling)}
\label{sec:data-LB}

A small portion of the entire data distribution, referred to as the trusted labeled dataset, is used to retrain a pre-trained Siamese neural network (SNN), known as \textit{``SiameseNet''}~\cite{melekhov2016siamese}. SNN is a neural network architecture consisting of two identical sub-networks (or branches) that share the same set of parameters. 

Let \( f_1(\cdot) \) and \( f_2(\cdot) \) represent the sub-networks within the Siamese network, sharing a common parameter \( \Theta \). Let \( f_{x_1} \) and \( f_{x_2} \) denote the final embedding vectors produced by the sub-networks, and \( h_{x_1} \) and \( h_{x_2} \) represent their output feature vectors, respectively. The trusted labeled dataset is used to train \textit{``SiameseNet''}. During training, each sub-network processes a different sample point at each iteration, extracts meaningful information, and generates feature embedding vectors. The final layer of \textit{``SiameseNet''} receives these embedding vectors, computes their statistical similarities, and outputs a similarity score. The final embedding feature vectors are represented as follows:

\begin{equation}\label{eq-2}
  \begin{split}
  h_{x_1} = f({x_c} ; \Theta) \\
  h_{x_2} = f({x_t} ; \Theta)
  \end{split}
\end{equation}

\noindent where $f({x_c})$ and $f({x_t})$ represent feature-embedded vectors with respect to each pair of the samples at every iteration, respectively.

Let \( x_p \) and \( x_q \) represent paired samples at each iteration. The network outputs \( 1 \) if it successfully captures the similarity between \( x_p \) and \( x_q \), and \( 0 \) otherwise. This learning process enables the \textit{``SiameseNet''} to distinguish between samples belonging to the same class and those that do not.

A "pairwise" contractive loss is incorporated into the learning process to encourage the network to produce identical embeddings for similar pairs and distinct embeddings for dissimilar pairs. This loss function penalizes the network when the statistical difference between identical pairs deviates significantly from a predefined threshold. The loss function is expressed as follows:

\begin{equation}\label{eq-loss}
  \begin{split}
    L_c(x_p, x_q, h) = (1-h) \cdot \frac{1}{2} (D_e(x_p, x_q))^2 + \\
    h \cdot \frac{1}{2}(max(0,\tau^*-D_e(x_p, x_q)))^2
  \end{split}
\end{equation}

In Equation~\ref{eq-loss}, $d(x_p, x_q)$ represents the Euclidean distance between $x_p$ and $x_q$, which equals $||f({x_c})-f({x_t})||$, where $L_c(x_p, x_q, h) = \frac{1}{2}(max(0,\tau^*-d(x_p, x_q)))^2$ when $h =1$, indicating high similarity between $x_p$ and $x_q$.

After training, the \textit{``SiameseNet''} is used to predict labels for clustered data points by computing their similarity scores with the small labeled dataset (trusted data distribution). During this process, a threshold \( \tau^* \) is introduced to evaluate the Euclidean distance (similarity score). This threshold determines whether a clustered sample is statistically similar to a labeled sample in the trusted distribution and should be assigned the same label. If the similarity score exceeds \( \tau^* \), the clustered sample is assigned the label of the most similar trusted sample. Otherwise, a unique label is assigned, indicating that the clustered sample does not belong to any class in the trusted distribution and is considered noise or an outlier. The optimal \( \tau^* \) is determined by balancing false positives (erroneously similar samples) and false negatives (missed genuine similarities). The selection of an appropriate \( \tau^* \) is based on precision, recall, and F1 score, which are defined as follows:

\begin{equation}\label{eq-4}
  Precision = \frac{TP}{TP + FP} \hspace{5mm} Recall =  \frac{TP}{TP + FN} \hspace{5mm}
  F1 = 2 \times \frac{Precision \times Recall}{Precision + Recall}
\end{equation}

\noindent where $TP$, $FP$, and $FN$ denote the true positive, false positive, and false negative, respectively. Then, we obtain an acceptable threshold $\tau^*$ through the maximization of the F1 score for various values of initial thresholds $\tau$, as expressed in Equation~\ref{eq-5}.

\begin{equation}\label{eq-5}
 \tau^* =  {argmax}_\tau \left(2 \times \frac{Precision(\tau) \times Recall(\tau)}{Precision(\tau) + Recall(\tau)}\right)
\end{equation}

To determine the optimal threshold \( \tau^* \) that maximizes the F1 score, we begin by defining a set of random \( \tau \) values. We then iteratively evaluate these values, computing the F1 score for each, and select the \( \tau \) that yields the highest F1 score.

To validate the accuracy of the labeling approach, a supervised cluster validation is performed by calculating the \textit{purity score}~\cite{reddy2013data} for each cluster. The purity score measures the extent to which a cluster contains data points with the same label, ranging from \( 0 \) to \( 1 \). A score of \( 1 \) indicates that all samples in the cluster belong to a single class, while a score of \( 0 \) suggests the opposite.

\subsection{Training Procedure for the MoE Framework}
\label{Sec:Tran-MoE-fram}
During each training phase, the goal is to maximize prediction accuracy/precision and minimize loss for both the expert models and the managing network. This requires simultaneous optimization of training hyperparameters for both models during forward and backward propagation.

An extensive hyperparameter optimization process is conducted for each expert model and the managing network. This includes determining the optimal learning rate, number of neurons per layer, batch size, number of convolutional filters, activation function type, and optimizer type. The specific hyperparameters to optimize depend on the task at hand.

The optimized hyperparameters, along with a cross-entropy loss approach, are used during joint training on each dataset. This ensures proper optimization of the objective functions (equations~\ref{eq-expert}~and~\ref{eq-gate}) for both the expert models and the managing network. Optimizing these functions minimizes predictive losses, enabling the experts to specialize effectively in their respective class distributions (clusters) and empowering the managing network to assign appropriate weights to each expert model. The objective functions combine cross-entropy loss with optimized parameters. For example, for the \textit{j-th} expert model, it can be expressed as follows:

\begin{equation}\label{eq-expert}
 L_{e_j} = -\sum_{i=1}^E log P(Y_i|X_i, \Theta_{e})
\end{equation}

In Equation~\ref{eq-expert}, $L_{e_j}$ is the objective function. \textit{S} denotes the total number of training samples per class, and $P(Y_i|X_i, \Theta_{e})$ represents the predictive probability for the actual (true) label $Y_i$ with respect to $X_i$ and $\Theta_{e}$. $X_i$ is considered to be from a specified class distribution assigned to \textit{j-th} expert model parameters. 

Furthermore, the objective function of the managing network can be expressed as follows:

\begin{equation}\label{eq-gate}
  L_g = -\sum_{i=1}^N log \sum_{j=1}^E g_j(X_i) P(Y_i|X_i, E_j, \Theta_g)
\end{equation}

\noindent where \textit{E} is the total number of expert models and $P(Y_i|X_i, E_j, \Theta_g)$ represents the predicted probability (output) from the \textit{j-th} expert model based on the assigned weights.

\textbf{\textit{(A) Forward Propagation:}} During the forward pass in the training phase, the managing network generates a set of weights \( g_j(X_i) \) corresponding to the number of expert models. Using these weights, the gating loss \( L_g \) is computed for the managing network. Each expert model produces an output \( P(Y_i|X_i, E_j) \) based on the gating weights \( g_j(X_i) \), and a loss \( L_{e_j} \) is calculated for each expert.

\textbf{\textit{(B) Loss Computation:}} For each input or set of inputs, the loss \( L_{e_j} \) of each expert model is computed by comparing the ground-truth label with the predicted combined output \( P(Y|X_i) \). If inputs from the same cluster are routed to different expert models, the managing network assigns different weights to these experts and combines their outputs. A combined loss is then calculated, as shown in Equation \ref{eq-com}, where \( Y \) represents the actual label.

\begin{equation}\label{eq-com}
  L_{com} = -\sum_{i=1}^N log \sum_{j=1}^K g_j(X_i) P(Y|X_i, E_j)
\end{equation}

\vspace{10pt}

\textbf{\textit{(C) Back Propagation:}}  
During the backward pass, the derivative of the objective function \( L_{e_j} \) with respect to the model's output \( P(Y|X_i, E_j) \) is calculated for a given input. This helps determine the gradient of the \textit{j-th} expert's loss relative to its output.

The computed gradient is then propagated backward through both the \textit{j-th} expert model and the managing network simultaneously. This process updates the weights of both models in a direction that maximizes their predictive accuracy and minimizes the loss for a given \( X_i \). The parameters \( \Theta_{e} \) of the \textit{j-th} expert model are updated based on their contribution to the overall output of the Mixture of Experts (MoE) model \( P(Y|X_i) \). Equation \ref{eq-E-para} provides the expression for calculating the gradient of \( L_{e_j} \) with respect to the parameter \( \Theta_{e} \) of the \textit{j-th} expert for a given \( X_i \).

\begin{equation}\label{eq-E-para}
  \frac{\partial L}  {\partial \Theta_{e}} = \sum_{i=1}^N g_j(X_i) \frac{\partial L_{ej}}{\partial  P(Y|X_i, E_j)}  \frac{\partial  P(Y|X_i, E_j)} {\partial \Theta_{e}}
\end{equation}

The gating function \( g_j(X_i) \) plays a key role in guiding the gradient for the \textit{j-th} expert, reflecting its contribution to the overall Mixture of Experts (MoE) output. The term \( \frac{\partial L_{ej}}{\partial P(Y|X_i, E_j)} \) represents how sensitive the expert's loss is to its predicted output, while \( \frac{\partial P(Y|X_i, E_j)}{\partial \Theta_{e}} \) calculates the gradient of the \textit{j-th} expert's output loss with respect to its parameters.

Hence, the gradient of  $L_{e_j}$ with respect to the \textit{j-th} expert model's output $P(Y|X_i, E_j)$ can be computed as follows (Equation~\ref{eq-E-output}):

\begin{equation}\label{eq-E-output}
  \frac{\partial L_{ek}}{\partial P(Y|X_i, E_j)} = g_j(X_i) \frac{1}{P(Y|X_i)} - 1_{[Y_i=j]}
\end{equation}

\noindent where $1_{[Y_i=j]}$ demonstrates an indicator function determining whether the actual label \textit{Y} corresponds to the  \textit{j-th} expert's class.

Similarly, for a given input \( X_i \), the gradient of the gating loss \( L_g \) is computed by taking its derivatives with respect to the gated weights \( g_j(X_i) \). The expression for calculating the gradient of the gating loss \( L_g \) with respect to the managing network's parameters \( \Theta_g \) for input \( X_i \) is provided in Equation \ref{eq-G-para}.

\begin{equation}\label{eq-G-para}
  \frac{\partial L}{\partial \Theta_{g}} = \sum_{i=1}^N  \sum_{j=1}^K  \frac{\partial L_g}{\partial g_j(X_i)}  \frac{\partial g_j(X_i)} {\partial \Theta_{g}}
\end{equation}

Whereas the gradient of the gated loss $L_g$ with respect to the gated output (weights) $g_j(X_i)$  given input $X_i$ and \textit{j-th} expert $E_j$, can be represented as follows:

\begin{equation}\label{eq-G-output}
  \frac{\partial L_g}{\partial g_j(X_i)} =  \frac{P(Y|X_i, E_j)}{\sum_{j=1}^K g_j(X_i) P(Y|X_i, E_j)}  - 1_{Y_i=j}
\end{equation}

The computed gradients from equations~\ref{eq-G-para} and \ref{eq-G-output}, along with the gradients obtained from the \textit{j-th} expert model in equations~\ref{eq-E-para} and \ref{eq-E-output}, are simultaneously propagated backward to the managing network. This guides the network in updating its weights. The gating parameters \( \Theta_g \) are updated based on the behavior of the gating loss \( L_g \) (i.e., whether the loss is decreasing or increasing in the output space). Effectively optimizing the objective function in Equation~\ref{eq-gate} and updating \( \Theta_g \) enables the managing network \( G \) to learn how to assign the correct weights \( g_j(X_i) \) to the appropriate expert for a given input.

The joint training process synchronizes weight updates for each expert and the managing network using the computed gradient information, ensuring that the overall loss of the Mixture of Experts (MoE) model is minimized for each input \( X_i \). Understanding the gradient characteristics is crucial for assessing and adjusting the learning dynamics of both the managing network and the expert models. This ensures adequate training and enables the model to comprehend the diversity in the input space, which is essential for its generation capability during inference.


\section{DFCP-MoE Framework Evaluation and Discussion}
\label{Sec:resuldis}

In this section, we present and discuss the results of our experimental evaluation, comparing the predictive performance of the DFCP-MoE framework with the traditional Mixture of Experts (MoE) and a dense model. Additionally, we highlight the significance of using the pre-trained Siamese neural network for data labeling. For our experiments, we utilize the GTSRB dataset (freely available at \url{https://benchmark.ini.rub.de/}) to validate the proposed framework.

\begin{table}[!h]
\caption{Selected optimized hyperparameters for the DFCP-MoE Expert Models based on the GTSRB using the Optuna optimization library.}
\label{tab:hyper}
\centering
\scriptsize
\begin{tabular}{l|c|cc|c|c|c|cc}
\toprule
\textbf{Experts} & \multicolumn{1}{c|}{\textbf{\begin{tabular}[c]{@{}c@{}}No. of \\ Training Samples\end{tabular}}} & \multicolumn{2}{c|}{\textbf{\begin{tabular}[c]{@{}c@{}}No. of\\ Neurons\end{tabular}}} & \multicolumn{1}{c|}{\begin{tabular}[c]{@{}c@{}}Learning\\ Rates\end{tabular}} & \textbf{Optimizers} & \begin{tabular}[c]{@{}c@{}}Batch\\ Size\end{tabular} & \multicolumn{2}{c}{\textbf{\begin{tabular}[c]{@{}c@{}}No. of Conv.\\ Filters\end{tabular}}} \\ \hline
                 &                                                                                                  & \multicolumn{1}{c|}{FC1}                   & \multicolumn{1}{c|}{FC2}                  &                                              &                     &                     & \multicolumn{1}{c|}{Conv1}                              & Conv2                              \\ 

\midrule
\hline

\textbf{1}       & 210                                                                                              & \multicolumn{1}{c|}{122}                   & 85                                        & 0.0040                                       & SGD                 & 16                  & \multicolumn{1}{c|}{6}                                  & 16                                 \\ \hline
\textbf{2}       & 2220                                                                                             & \multicolumn{1}{c|}{142}                   & 95                                        & 0.0019                                       & RMSPro              & 64                  & \multicolumn{1}{c|}{11}                                 & 22                                 \\ \hline
\textbf{3}       & 2250                                                                                             & \multicolumn{1}{c|}{142}                   & 95                                        & 0.0018                                       & RMSPro              & 64                  & \multicolumn{1}{c|}{11}                                 & 22                                 \\ \hline
\textbf{4}       & 1410                                                                                             & \multicolumn{1}{c|}{134}                   & 91                                        & 0.0029                                       & Adam                & 64                  & \multicolumn{1}{c|}{14}                                 & 29                                 \\ \hline
\textbf{5}       & 1980                                                                                             & \multicolumn{1}{c|}{139}                   & 93                                        & 0.0025                                       & RMSPro              & 64                  & \multicolumn{1}{c|}{14}                                 & 28                                 \\ \hline
\textbf{6}       & 1860                                                                                             & \multicolumn{1}{c|}{138}                   & 93                                        & 0.0025                                       & RMSPro              & 64                  & \multicolumn{1}{c|}{10}                                 & 20                                 \\ \hline
\textbf{7}       & 420                                                                                              & \multicolumn{1}{c|}{124}                   & 86                                        & 0.0037                                       & SGD                 & 32                  & \multicolumn{1}{c|}{8}                                  & 17                                 \\ \hline
\textbf{8}       & 1440                                                                                             & \multicolumn{1}{c|}{134}                   & 91                                        & 0.0029                                       & Adam                & 64                  & \multicolumn{1}{c|}{13}                                 & 32                                 \\ \hline
\textbf{9}       & 1410                                                                                             & \multicolumn{1}{c|}{134}                   & 91                                        & 0.0029                                       & Adam                & 64                  & \multicolumn{1}{c|}{14}                                 & 29                                 \\ \hline
\textbf{10}      & 1470                                                                                             & \multicolumn{1}{c|}{135}                   & 91                                        & 0.0030                                       & Adam                & 64                  & \multicolumn{1}{c|}{14}                                 & 29                                 \\ \hline
\textbf{11}      & 2010                                                                                             & \multicolumn{1}{c|}{140}                   & 94                                        & 0.0025                                       & RMSPro              & 64                  & \multicolumn{1}{c|}{12}                                 & 24                                 \\ \hline
\textbf{12}      & 1320                                                                                             & \multicolumn{1}{c|}{133}                   & 90                                        & 0.0032                                       & Adam                & 64                  & \multicolumn{1}{c|}{13}                                 & 26                                 \\ \hline
\textbf{13}      & 2100                                                                                             & \multicolumn{1}{c|}{141}                   & 94                                        & 0.0023                                       & RMSPro              & 64                  & \multicolumn{1}{c|}{11}                                 & 22                                 \\ \hline
\textbf{14}      & 2160                                                                                             & \multicolumn{1}{c|}{141}                   & 94                                        & 0.0020                                       & RMSPro              & 64                  & \multicolumn{1}{c|}{11}                                 & 22                                 \\ \hline
\textbf{15}      & 780                                                                                              & \multicolumn{1}{c|}{127}                   & 88                                        & 0.0038                                       & Adam                & 64                  & \multicolumn{1}{c|}{12}                                 & 24                                 \\ \hline
\textbf{16}      & 630                                                                                              & \multicolumn{1}{c|}{126}                   & 87                                        & 0.0034                                       & Adam                & 32                  & \multicolumn{1}{c|}{9}                                  & 19                                 \\ \hline
\textbf{17}      & 420                                                                                              & \multicolumn{1}{c|}{124}                   & 86                                        & 0.0037                                       & SGD                 & 32                  & \multicolumn{1}{c|}{8}                                  & 17                                 \\ \hline
\textbf{18}      & 1110                                                                                             & \multicolumn{1}{c|}{131}                   & 89                                        & 0.0024                                       & Adam                & 64                  & \multicolumn{1}{c|}{13}                                 & 26                                 \\ \hline
\textbf{19}      & 1200                                                                                             & \multicolumn{1}{c|}{132}                   & 90                                        & 0.0022                                       & Adam                & 64                  & \multicolumn{1}{c|}{12}                                 & 24                                 \\ \hline
\textbf{20}      & 210                                                                                              & \multicolumn{1}{c|}{122}                   & 85                                        & 0.0040                                       & SGD                 & 16                  & \multicolumn{1}{c|}{6}                                  & 16                                 \\ \hline
\textbf{21}      & 360                                                                                              & \multicolumn{1}{c|}{123}                   & 85                                        & 0.0038                                       & SGD                 & 16                  & \multicolumn{1}{c|}{7}                                  & 18                                 \\ \hline
\textbf{22}      & 330                                                                                              & \multicolumn{1}{c|}{123}                   & 85                                        & 0.0034                                       & SGD                 & 16                  & \multicolumn{1}{c|}{6}                                  & 16                                 \\ \hline
\textbf{23}      & 390                                                                                              & \multicolumn{1}{c|}{123}                   & 85                                        & 0.0036                                       & SGD                 & 16                  & \multicolumn{1}{c|}{7}                                  & 18                                 \\ \hline
\textbf{24}      & 510                                                                                              & \multicolumn{1}{c|}{125}                   & 86                                        & 0.0034                                       & Adam                & 32                  & \multicolumn{1}{c|}{18}                                 & 17                                 \\ \hline
\textbf{25}      & 270                                                                                              & \multicolumn{1}{c|}{122}                   & 85                                        & 0.0039                                       & SGD                 & 16                  & \multicolumn{1}{c|}{6}                                  & 16                                 \\ \hline
\textbf{26}      & 1500                                                                                             & \multicolumn{1}{c|}{135}                   & 91                                        & 0.0027                                       & Adam                & 64                  & \multicolumn{1}{c|}{15}                                 & 30                                 \\ \hline
\textbf{27}      & 600                                                                                              & \multicolumn{1}{c|}{126}                   & 87                                        & 0.0022                                       & Adam                & 32                  & \multicolumn{1}{c|}{9}                                  & 19                                 \\ \hline
\textbf{28}      & 240                                                                                              & \multicolumn{1}{c|}{122}                   & 85                                        & 0.0040                                       & SGD                 & 16                  & \multicolumn{1}{c|}{6}                                  & 16                                 \\ \hline
\textbf{29}      & 540                                                                                              & \multicolumn{1}{c|}{125}                   & 86                                        & 0.0034                                       & Adam                & 32                  & \multicolumn{1}{c|}{9}                                  & 18                                 \\ \hline
\textbf{30}      & 270                                                                                              & \multicolumn{1}{c|}{122}                   & 84                                        & 0.0039                                       & SGD                 & 16                  & \multicolumn{1}{c|}{6}                                  & 16                                 \\ \hline
\textbf{31}      & 450                                                                                              & \multicolumn{1}{c|}{124}                   & 86                                        & 0.0036                                       & SGD                 & 32                  & \multicolumn{1}{c|}{8}                                  & 17                                 \\ \hline
\textbf{32}      & 780                                                                                              & \multicolumn{1}{c|}{127}                   & 88                                        & 0.0031                                       & Adam                & 64                  & \multicolumn{1}{c|}{12}                                 & 24                                 \\ \hline
\textbf{33}      & 240                                                                                              & \multicolumn{1}{c|}{122}                   & 85                                        & 0.0040                                       & SGD                 & 16                  & \multicolumn{1}{c|}{6}                                  & 16                                 \\ \hline
\textbf{34}      & 689                                                                                              & \multicolumn{1}{c|}{126}                   & 87                                        & 0.0033                                       & Adam                & 32                  & \multicolumn{1}{c|}{9}                                  & 19                                 \\ \hline
\textbf{35}      & 420                                                                                              & \multicolumn{1}{c|}{124}                   & 86                                        & 0.0037                                       & SGD                 & 32                  & \multicolumn{1}{c|}{8}                                  & 17                                 \\ \hline
\textbf{36}      & 1200                                                                                             & \multicolumn{1}{c|}{134}                   & 90                                        & 0.0020                                       & Adam                & 64                  & \multicolumn{1}{c|}{12}                                 & 24                                 \\ \hline
\textbf{37}      & 390                                                                                              & \multicolumn{1}{c|}{123}                   & 85                                        & 0.0038                                       & SGD                 & 16                  & \multicolumn{1}{c|}{7}                                  & 18                                 \\ \hline
\textbf{38}      & 210                                                                                              & \multicolumn{1}{c|}{122}                   & 85                                        & 0.0040                                       & SGD                 & 16                  & \multicolumn{1}{c|}{6}                                  & 16                                 \\ \hline
\textbf{39}      & 2070                                                                                             & \multicolumn{1}{c|}{140}                   & 94                                        & 0.0022                                       & RMSPro              & 64                  & \multicolumn{1}{c|}{11}                                 & 22                                 \\ \hline
\textbf{40}      & 300                                                                                              & \multicolumn{1}{c|}{122}                   & 85                                        & 0.0038                                       & SGD                 & 16                  & \multicolumn{1}{c|}{6}                                  & 16                                 \\ \hline
\textbf{41}      & 360                                                                                              & \multicolumn{1}{c|}{123}                   & 85                                        & 0.0047                                       & SGD                 & 16                  & \multicolumn{1}{c|}{7}                                  & 18                                 \\ \hline
\textbf{42}      & 240                                                                                              & \multicolumn{1}{c|}{122}                   & 85                                        & 0.0040                                       & SGD                 & 16                  & \multicolumn{1}{c|}{6}                                  & 16                                 \\ \hline
\textbf{43}      & 240                                                                                              & \multicolumn{1}{c|}{122}                   & 85                                        & 0.0040                                       & SGD                 & 16                  & \multicolumn{1}{c|}{6}                                  & 16                                 \\ 
\bottomrule
\end{tabular}
\end{table}

\begin{table}[!h]
\caption{Optimized parameters and values for the dense model and the managing neural network.}
\label{tab:sinMan}
\centering
    \begin{tabular}{l|c|c}
    \toprule
    \textbf{Tuned Hyperparameters} & \textbf{Dense model} & \textbf{Managing network} \\
    \midrule
    \hline

    Learning rate          & 0.001          & 0.0001        \\ \hline

    Optimizer              & Adam           & Adam          \\ \hline
    
    Batch size             & 32             & 32            \\ \hline
    
    Number of Epochs      & 100             & 100           \\ \hline
    
    Conv1 (No. of Conv. filters)            & 32       & 64             \\ \hline
    Conv2 (No. of Conv. filters)           & 16        & 128            \\ \hline
    Conv3  (No. of Conv. filters)           & 24         & 256             \\ \hline
    
    Conv4 (No. of Conv. filters)      & 40          & {-}             \\ \hline
    
    Conv5 (No. of Conv. filters)      & 80             & {-}              \\ \hline
    
    Conv6 (No. of Conv. filters)     & 112             & {-}               \\ \hline
    
    Conv7 (No. of Conv. filters)     & 192          & {-}               \\ \hline
    
    No. of neurons (fully-connected layer) & 1280  & (FC1) = 512 (FC2) =128   \\ \hline
    
    Pooling   & AdaptiveAvgPool2d   & Maxpooling                \\ \hline
    
    Dropout                  & 0.6          & 0.5                       \\ \hline
    
    Weight initialization     & Pretrained Weights   & Xavier Normal    \\ \hline
    
    Activation Function         & ReLu           & ReLu                   \\ \hline
    
    Learning Rate Scheduler & StepLR & ReduceLROnPlateau  \\ \hline
    
    Weight Decay   & 1e-4    & 1e-4                      \\ \hline
    
    Loss Function    & CrossEntropy    & CrossEntropy    \\ \hline
    
    Number of Training Samples   & 39,209    & 39,209     \\ \hline
      
         \bottomrule
    \end{tabular}
\end{table}





















In this context, the traditional MoE refers to an MoE model trained without clustering the training data or conditioning the managing network to assign specific weights to specific experts, as commonly done in previous works~\cite{ma2018modeling, yuksel2012twenty}. A dense model, on the other hand, refers to a deep neural network (DNN) trained on the entire dataset using standard deep learning techniques. These experiments aim to determine whether the DFCP-MoE model outperforms the traditional MoE model and the dense model for an image classification task on the GTSRB dataset. 

For these experiments, a LeNet-5 image classification model is used for both the expert models and the managing network in both the DFCP-MoE framework and the traditional MoE model. For the dense model, an EfficientNet-B1 transfer learning model is employed. It is important to note that the original parameters of these architectures are adjusted during training due to hyperparameter tuning, as detailed in Tables~\ref{tab:hyper}~and~\ref{tab:sinMan}.

\begin{figure}[!h]
    \centering
    \includegraphics[width=\textwidth]{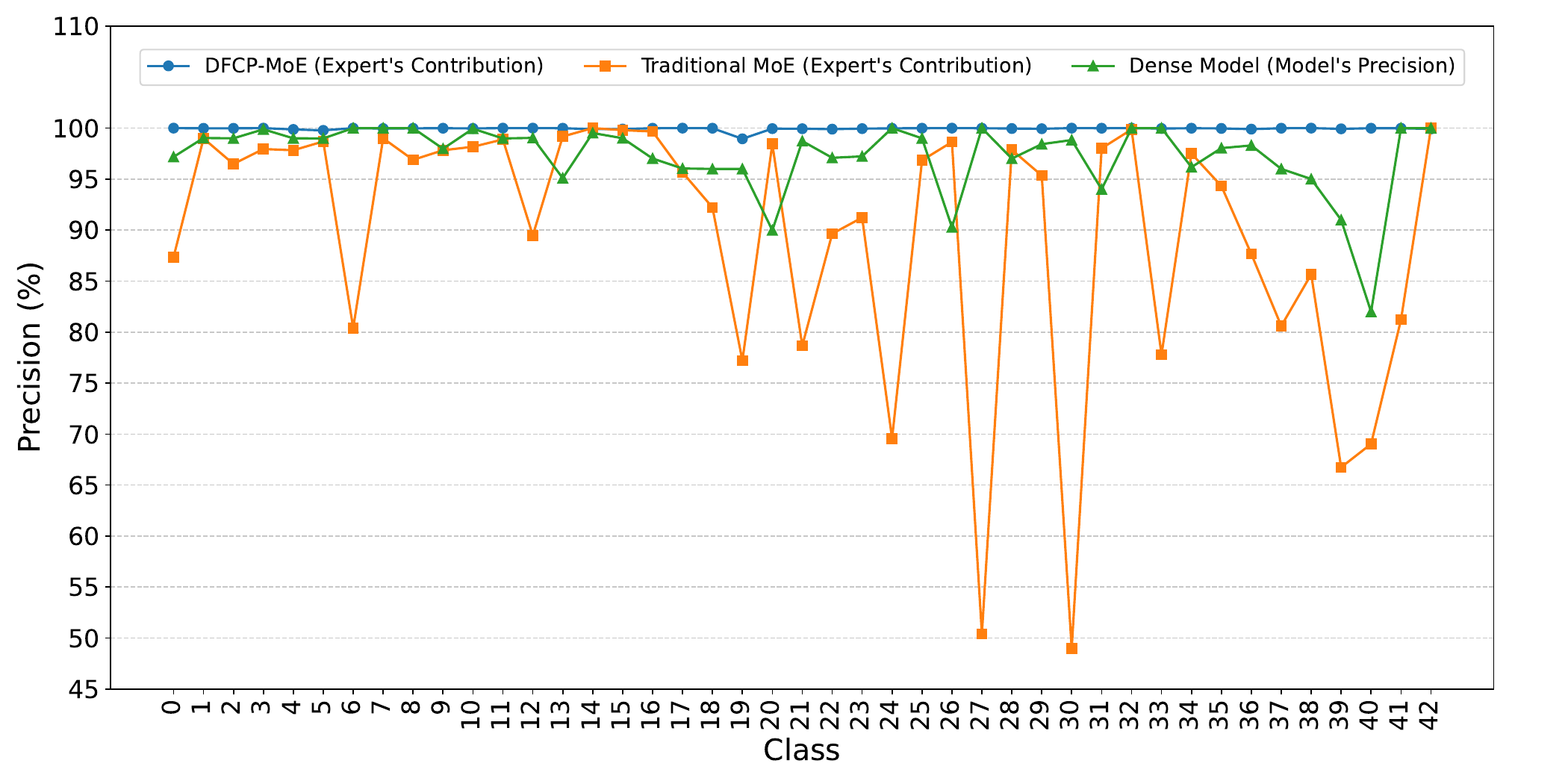}
    \caption{Evaluation plot for performance comparison between the proposed DFCP-MoE, a traditional MoE, and a dense model over the GTSRB dataset.}
    \label{fig:eval_1}
\end{figure}

Our experimental results in Figure~\ref{fig:eval_1} demonstrate that over \( 95\% \) of the experts within the DFCP-MoE framework consistently make significant contributions to their assigned class distributions. This confirms the success of our feature similarity checking and data labeling processes. The feature similarity checking is performed by the \textit{``SiameseNet''}, a well-optimized, pre-trained model originally trained on a large dataset. Leveraging transfer learning principles, it can be reused even with small datasets for related tasks. The model learns to pair samples based on their statistical similarities, where similar pairs belong to the same class and dissimilar pairs belong to different classes.

Following the DFCP-MoE framework specifications, we compute the purity score for each cluster after the pseudo-labeling stage (see Section~\ref{sec:data-LB}). This score acts as a cluster validation metric, indicating whether all samples in a cluster belong to the same class. We achieved an average cluster purity score \( 0.98 \), highlighting the accuracy of the double-state clustering technique in grouping samples of the same class distribution. After cluster purity evaluations, we obtained \( 43 \) clusters and, consequently, \( 43 \) expert models for both the proposed DFCP-MoE and traditional MoE frameworks.

In the training setup for the DFCP-MoE framework, the traditional MoE model, and the dense model, we incorporate a hyperparameter tuning technique using the PyTorch Optuna framework. This helps optimize training hyperparameters, such as the number of neurons, learning rate, batch size, and optimizer, for the expert models, managing networks, and dense model. During this process, optimal hyperparameters are selected for each model based on their respective number of training samples.

Since each expert in the DFCP-MoE framework is trained on data from a specific class distribution (a specific cluster) with a unique number of training samples, the hyperparameter optimization algorithm assigns distinct hyperparameters to each expert. Table~\ref{tab:hyper} displays the selected training hyperparameters and the number of training samples for each expert model within the DFCP-MoE framework. Meanwhile, Table~\ref{tab:sinMan} presents the training hyperparameters and the number of training samples for the dense model and the managing network.

Selecting ideal hyperparameters for the DFCP-MoE and traditional MoE frameworks refines expert assignment and gating operations, leading to enhanced precision and improved overall model performance. Additionally, optimal hyperparameters ensure efficient use of computational resources by balancing the contributions of each expert to the MoE model's final output, reducing redundancy, and improving both inference and training efficiency. For dense models, choosing the best hyperparameters helps prevent overfitting and vanishing gradients, accelerating the loss convergence process.

Table~\ref{tab:hyper} shows that the optimization algorithm selects a higher number of neurons and a lower learning rate for expert models with a large number of training samples. This enables the experts to effectively capture the intricate details and complexities of the inputs within their assigned clusters. Conversely, for expert models with fewer training samples, a smaller number of neurons and a higher learning rate are chosen to avoid overfitting and facilitate optimal learning from limited data. Notably, similar hyperparameters are selected for experts in both the DFCP-MoE framework and the traditional MoE, as experts in both frameworks have training samples within the same numerical range.

Our experiments highlight the importance of using different optimizers and batch sizes for expert models based on the number of training samples in their respective class distributions. This approach allows for optimal fine-tuning of each expert model according to its specific class distribution, improving its inference predictive performance and contribution to the MoE's final output. Optimization strategies such as SGD, Adam, and RMSprop offer distinct advantages and perform differently depending on the characteristics and size of the training data.

Additionally, the hyperparameter optimization algorithm selects smaller batch sizes for experts with fewer training samples (see Table~\ref{tab:hyper}). This enables more frequent updates of the expert model's weights during training, enhancing its generalization during inference. Conversely, for expert models with a larger number of training samples, the algorithm selects larger batch sizes, which improves training and inference consistency and efficiency.

The number of filters in each convolutional layer for an expert model determines the dimensionality of its output feature maps, influencing its ability to learn intricate patterns in the input space. As shown in Table~\ref{tab:hyper}, each expert is assigned a varying number of convolutional filters. This allows each expert to extract features at different levels of granularity, enabling them to capture unique attributes of data samples within their respective class distributions.

The performance of each model is evaluated using metrics such as ``Mean Average Precision (mAP)'', ``Balanced Accuracy``, ``Average Inference Time'', and ``Training Duration''. These metrics were chosen because they provide a comprehensive assessment of classification performance, efficiency, and practical relevance. Each metric highlights a specific aspect of the models' potential, offering a balanced perspective on the trade-offs between predictive accuracy and efficiency, which are critical for this study. Additionally, we assess the classification performance of each model by examining the precision of each expert's contribution to the final output of the DFCP-MoE and traditional MoE frameworks, as well as the precision of the dense model.

To facilitate a thorough comparison of classification performance, we present our findings in both tabular and graphical formats. This dual approach allows for a comprehensive and intuitive understanding of the results, emphasizing key conclusions that may not be immediately apparent from tables or graphs alone. The tabular presentation (Table \ref{tab:ad_eval_1}) concisely summarizes the key evaluation metrics for all three models, highlighting their performance and potential for computational simplicity, particularly during inference.

\begin{table}[!h]
\caption{Selected evaluation metrics for performance comparison between the proposed DFCP-MoE, a traditional MoE, and a dense model for GTSRB image classification.}
\label{tab:ad_eval_1}
\centering
\begin{tabular}{l|c|c|c}
\toprule

\textbf{Metrics} & \textbf{DFCP-MoE} & \textbf{\begin{tabular}[c]{@{}c@{}}Traditional\\    MoE\end{tabular}} & \textbf{\begin{tabular}[c]{@{}c@{}}Dense \\ Model\end{tabular}} \\ \hline \hline

Mean Average Precision (mAP)  & 99.95\%  & 92.67\%                    & 96.68\%               \\ \hline

Balanced Accuracy  & 0.99   & 0.98      & 0.99                  \\ \hline

Average Inference Time  & $\sim$51ms   & $\sim$50ms           & $\sim$71ms            \\ \hline

Training Time  & $\sim$126 Hours     & $\sim$129 Hours       & $\sim$3.48 Hours      \\ \hline

\bottomrule
\end{tabular}

\end{table}

The ``mean average precision (mAP)'' measures the model's ability to distinguish relevant information from irrelevant information. A higher mAP score indicates better precision-recall trade-offs, reflecting improved accuracy in identifying key features in the input and enhancing inference performance. ``Balanced accuracy'' evaluates how well the model performs across all classes during training. A higher score suggests the model is less biased, even with an imbalanced dataset, while a lower score indicates the opposite. This metric is typically measured during inference.

Table~\ref{tab:ad_eval_1} shows that our proposed DFCP-MoE outperforms both the traditional MoE and the dense model, achieving superior mAP and balanced accuracy scores. ``Average inference time'' measures the mean duration for a model to make a prediction on a single input instance during inference. Alongside precision, this metric is crucial for assessing a model's efficiency and real-world applicability. As seen in Table~\ref{tab:ad_eval_1}, the DFCP-MoE exhibits the lowest average inference time compared to the traditional MoE and dense model, demonstrating its efficiency and suitability for real-time applications. This reduced inference time results from the managing network's optimal weight allocation to experts, eliminating unnecessary computations while maintaining high precision. This efficiency is vital for low-latency applications such as autonomous driving systems, face recognition, and biometric finger scanning, where higher inference times could hinder real-world usability.

Generally, a dense model produces a straightforward predictive output compared to both the DFCP-MoE and traditional MoE models. In an MoE framework, the managing network must receive input and determine which expert to assign it to for prediction, which can introduce additional latency during inference. However, in our experiments, the dense model's architecture is inherently more complex than both the managing network and the expert models due to the dataset's characteristics. Additionally, the dense model exhibits higher latency compared to the two MoE frameworks because, in practice, all its parameters are activated for every input, increasing inference time. 

We designed the evaluation phase for the DFCP-MoE and traditional MoE such that the initial layers of all experts are pre-computed in parallel with the managing network. This optimization significantly reduces the inference time for the MoE models, allowing them to process and make predictive inferences faster than the dense model.

As shown in Table~\ref{tab:ad_eval_1}, the **training time** for the DFCP-MoE and traditional MoE models is significantly higher than that of the dense model. This is because backpropagation is simultaneously coordinated during the joint training of the managing network and each expert model for every epoch, which increases computational time. Additionally, optimizing hyperparameters for both the expert models and the managing network further extends the training time for the MoE models. The overall training duration for the MoE models is calculated by summing the training durations of all expert models and the managing network, resulting in a much longer training period compared to the dense model.

However, the training duration of the traditional MoE is longer than that of the DFCP-MoE model. In the traditional MoE, the managing network randomly assigns weights to experts, requiring more time to determine which input is best suited for which expert. In contrast, the DFCP-MoE's managing network is preconditioned to assign specific weights to specific experts based on cluster assignments, reducing the time needed for weight allocation.

The graphical presentation in Figure \ref{fig:eval_1} provides additional classification performance assessments for each model, highlighting their predictive ability across class distributions in the dataset. The graph demonstrates that the proposed DFCP-MoE framework consistently outperforms both the traditional MoE and the dense model in terms of inference-based classification precision for most classes. This indicates that the conditional end-to-end training of the DFCP-MoE framework enables experts to develop more specialized knowledge, resulting in higher predictive precision for their respective classes compared to the traditional MoE and the dense model. The average contribution of each expert reflects the average precision of predictions for inputs within their assigned class distributions. Since each expert is trained on a specific class, the class notations on the \textit{x-axis} of each plot represent specialized experts.


 \section{Conclusions and Future Work}
\label{Sec:Conlfuwork}
In this work, we introduced the \textit{Double-stage Feature-level Clustering and Pseudo-labeling-based Mixture of Experts (DFCP-MoE)} framework, which advances the mixture of experts technique for image classification tasks. The DFCP-MoE framework employs multiple DNNs as expert models and a gate network to address complex machine learning problems and enhance the reliability of their outputs. We trained, evaluated, and compared the training and inference performance of the DFCP-MoE model with a traditional MoE model and a dense model on the GTSRB dataset for classification tasks. The DFCP-MoE model provides a robust approach for handling classification tasks by utilizing well-specialized experts for different class distributions and maintaining an optimized managing network to control and coordinate the experts. All three approaches are supported by a hyperparameter optimization technique, allowing the use of distinct hyperparameters for each expert model based on the number of training samples. The evaluation results demonstrate that the DFCP-MoE model outperforms both the traditional MoE and the dense model, validating the effectiveness of clustering the input space into subdomains and using multiple models for training and inference, despite the increased computational complexity during training. Future research could explore other optimization methods and adapt the DFCP-MoE framework to different data types, such as video datasets. Additionally, outliers in the data may impact the precision of the two-phase clustering and data labeling stages. Therefore, developing a robust statistical methodology to identify and evaluate the legitimacy of these outliers presents another promising research direction.

\section{Acknowledgments}
\label{Sec:acknow}

This work was supported by the Fundação para a Ciência e a Tecnologia (FCT) under Ph.D. grant agreement No 2023.02832.BD through the LASIGE Research Unit (ref. UID/000408/2025).


\section*{Acknowledgments}
This work was supported by the LASIGE Research Unit (ref. UIDB/00408/2020 and ref. UIDP/00408/2020).

\end{document}